\pdfoutput=1

\documentclass[11pt]{article}

\usepackage[]{acl}

\usepackage{times}
\usepackage{latexsym}
\usepackage{graphicx}
\usepackage{amsmath,amsthm,bm}
\usepackage{amsfonts}
\usepackage{booktabs}
\usepackage{multirow}
\usepackage{color}
\usepackage{subfigure}
\usepackage{CJK}
\usepackage{tcolorbox}
\usepackage{tabu}
\usepackage{dashrule}

\tcbuselibrary{theorems}
\tcbset{highlight math/.append style={left=0mm,right=0mm,top=0mm,bottom=0mm, colframe=white}}
\newtcbox{\mybox}[1][red]
  {on line, arc = 0pt, outer arc = 0pt,
    colback = #1!10!white, colframe = #1!50!black,
    boxsep = 0pt, left = 1pt, right = 1pt, top = 2pt, bottom = 2pt,
    boxrule = 0pt}

\newtcbox{\wrongbox}[1][blue]
  {on line, arc = 0pt, outer arc = 0pt,
    colback = #1!10!white, colframe = #1!50!black,
    boxsep = 0pt, left = 1pt, right = 1pt, top = 2pt, bottom = 2pt,
    boxrule = 0pt}

\usepackage[T1]{fontenc}

\usepackage[utf8]{inputenc}

\usepackage{microtype}

\title{Unsupervised Boundary-Aware Language Model Pretraining for Chinese Sequence Labeling}

\author{
  Peijie Jiang$^1$ ~~~~ Dingkun Long ~~~~ Yanzhao Zhang ~~~~ Pengjun Xie \\
  {\bf Meishan Zhang$^2\thanks{~~Corresponding author.}$ ~~~~ Min Zhang$^2$} \\
  $^1$School of New Media and Communication, Tianjin University, China\\
  $^2$Institute of Computing and Intelligence, Harbin Institute of Technology (Shenzhen)\\
  \texttt{jzx555@tju.edu.cn}, \texttt{\{zhangmeishan,zhangmin2021\}@hit.edu.cn}\\
  \texttt{\{longdingkun1993,zhangyanzhao00,xpjandy\}@gmail.com}\\
}

\begin{document}
\begin{CJK}{UTF8}{gbsn}

\maketitle
\begin{abstract}
  Boundary information is critical for various Chinese language processing tasks, such as word segmentation, part-of-speech tagging, and named entity recognition.
  Previous studies usually resorted to the use of a high-quality external lexicon, where lexicon items can offer explicit boundary information. 
  However, to ensure the quality of the lexicon, great human effort is always necessary, which has been generally ignored.  
  In this work, we suggest unsupervised statistical boundary information instead,   
  and propose an architecture to encode the information directly into pre-trained language models,
  resulting in Boundary-Aware BERT (BABERT).
  We apply BABERT for feature induction of Chinese sequence labeling tasks. 
  Experimental results on ten benchmarks of Chinese sequence labeling demonstrate that BABERT can provide consistent improvements on all datasets.
  In addition,
  our method can complement previous supervised lexicon exploration,
  where further improvements can be achieved when integrated with external lexicon information.
  
\end{abstract}

\section{Introduction}
The representative sequence labeling tasks for the Chinese language, such as word segmentation,
part-of-speech (POS) tagging and named entity recognition (NER) \cite{emerson-2005-second, jin-chen-2008-fourth},
have been inclined to be performed at the character-level in an end-to-end manner \cite{shen-etal-2016-consistent}.
The paradigm, naturally, is standard to Chinese word segmentation (CWS),
while for Chinese POS tagging and NER,
it can better help reduce the error propagation \cite{sun-uszkoreit-2012-capturing, Yang2016Combining, liu-etal-2019-encoding}
compared with word-based counterparts by straightforward modeling.


Recently, all the above tasks have reached state-of-the-art performances with the help of BERT-like pre-trained language models \cite{yan2019tener, meng2019glyce}.
The BERT variants, such as BERT-wwm \cite{cui2021pre}, ERNIE \cite{sun2019ernie}, ZEN \cite{diao-etal-2020-zen}, NEZHA \cite{wei2019nezha}, etc.,
further improve the vanilla BERT by either using external knowledge or larger-scale training corpus.
The improvements can also benefit character-level Chinese sequence labeling tasks.


Notably, since the output tags of all these character-level Chinese sequence labeling tasks involve 
identifying Chinese words or entities \cite{zhang-yang-2018-chinese, yang-etal-2019-subword},
prior boundary knowledge could be highly helpful for them.
A number of studies propose the integration of an external lexicon to enhance their baseline models
by feature representation learning \cite{jia-etal-2020-entity, tian-etal-2020-joint-chinese, liu-etal-2021-lexicon}.
Moreover, some works suggest injecting similar resources into the pre-trained BERT weights.
BERT-wwm \cite{cui2021pre} and ERNIE \cite{sun2019ernie} are the representatives,
which leverage an external lexicon for masked word prediction in Chinese BERT.

The lexicon-based methods have indeed achieved great success for boundary integration.
However, there are two major drawbacks.
First, the lexicon resources are always constructed manually \cite{zhang-yang-2018-chinese, diao-etal-2020-zen, jia-etal-2020-entity, liu-etal-2021-lexicon}, which is expensive and time-consuming.
The quality of the lexicon is critical to our tasks.
Second, different tasks as well as different domains require different lexicons \cite{jia-etal-2020-entity, liu-etal-2021-lexicon}.
A well-studied lexicon for word segmentation might be inappropriate for NER,
and a lexicon for news NER might also be problematic for finance NER.
The two drawbacks can be due to the supervised characteristic of these lexicon-based enhancements.
Thus, it is more desirable to offer boundary information in an unsupervised manner.



In this paper, we propose an unsupervised Boundary-Aware BERT (BABERT),
which is achieved by fully exploring the potential of statistical features mined from a large-scale raw corpus.
We extract a set of N-grams (a predefined fixed N) no matter they are valid words or entities,
and then calculate their corresponding unsupervised statistical features, which are mostly related to boundary information.
We inject the boundary information into the internal layer of a pre-trained BERT,
so that our final BABERT model can approximate the boundary knowledge softly by using inside representations.
The BABERT model has no difference from the original BERT,
so that we can use it in the same way as the standard BERT exploration.


We conduct experiments on three Chinese sequence labeling tasks to demonstrate the effectiveness of our proposed method.
Experimental results show that our approach can significantly outperform other Chinese pre-trained language models.
In addition, compared with supervised lexicon-based methods, BABERT obtains competitive results on all tasks and achieves further improvements when integrated with external lexicon knowledge.
We also conduct extensive analyses to understand our method comprehensively\footnote{The pre-trained model and code will be publicly available at \url{http://github.com/modelscope/adaseq/examples/babert}}.

Our contributions in this paper include the following:
1) We design a method to encode unsupervised statistical boundary information into boundary-aware representation,
2) propose a new pre-trained language model called BABERT as a boundary-aware extension for BERT,
3) verify BABERT on ten benchmark datasets of three Chinese sequence labeling tasks.









\begin{figure*}
  \centering
  \includegraphics[scale=0.48]{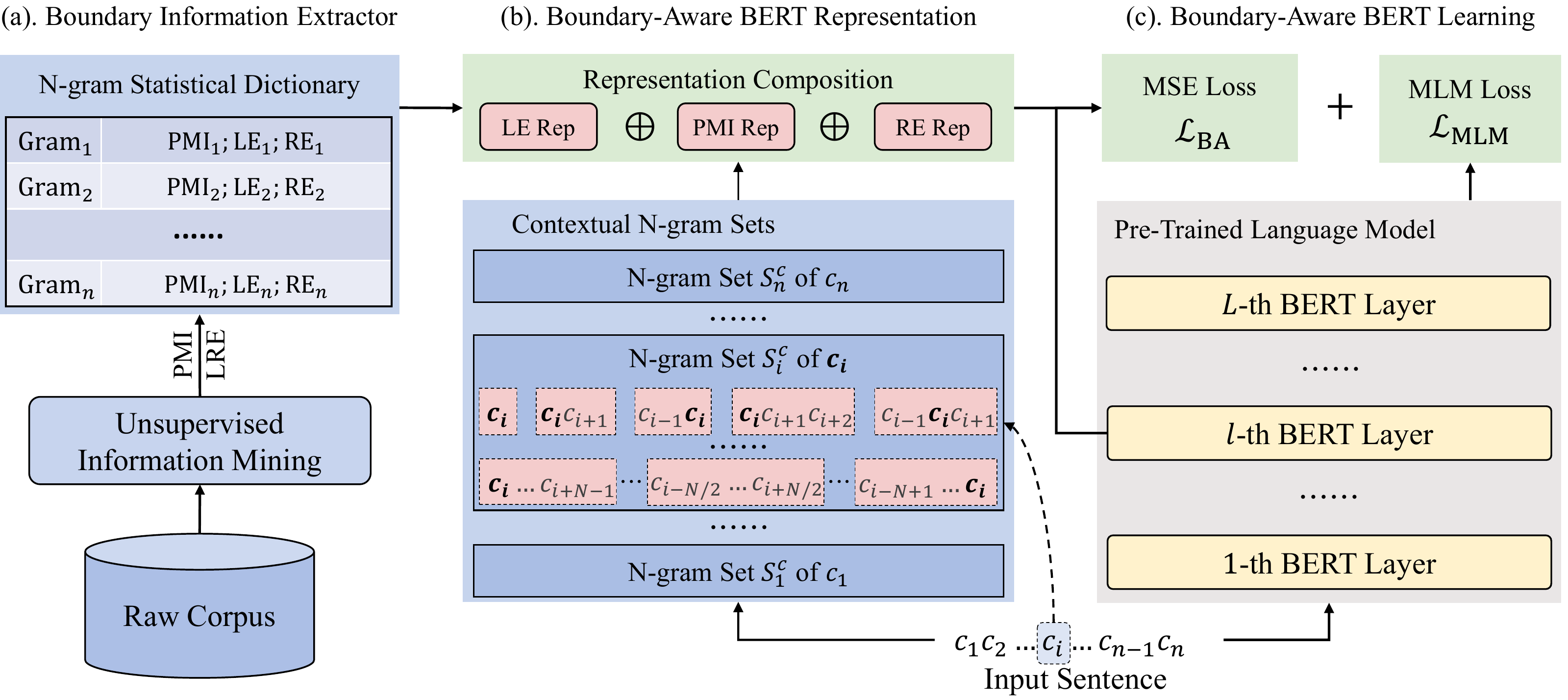}
  \caption{The overall architecture of the boundary-aware pre-trained language model, which consists of three parts:
  (a) boundary information extractor, 
  (b) boundary-aware representation, 
  and (c) boundary-aware BERT Learning. 
  The boundary-aware objective $\mathcal{L}_\texttt{BA}$ is defined in Equation \ref{eq:BA}.
  }
  \label{fig:model}
\end{figure*}

\section{Related Work}
In the past decades, machine learning has achieved good performance on sequence labeling tasks with statistical information \cite{bellegarda2004statistical, low2005maximum, bouma2009normalized}.
Recently, neural models have led to state-of-the-art results for Chinese sequence labeling \cite{lample-etal-2016-neural, ma-hovy-2016-end, TACL792}.
In addition, the presence of language representation models such as BERT \cite{devlin2018bert} has led to impressive improvements.
In particular, many variants of BERT are devoted to integrating boundary information into BERT to improve Chinese sequence labeling \cite{diao-etal-2020-zen,jia-etal-2020-entity,liu-etal-2021-lexicon}.

\paragraph{Statistical Machine Learning}
Statistical information is critical for sequence labeling.
Previous works attempt to count such information from large corpora in order to combine it with machine learning methods for sequence labeling \cite{bellegarda2004statistical, liang2005semi, bouma2009normalized}.
\citet{peng2004chinese} attempts to conduct sequence labeling by CRF and a statistical-based new word discovery method.
\citet{low2005maximum} introduce a maximum entropy approach for sequence labeling.
\citet{liang2005semi} utilizes unsupervised statistical information in Markov models,
and gets a boost on Chinese NER and CWS.

\paragraph{Pre-trained Language Model}
Pre-trained language model is a hot topic in natural language processing (NLP) communities \cite{devlin2018bert, liu2019roberta, wei2019nezha, clark2020electra, diao-etal-2020-zen, zhang-etal-2021-ambert}
and has been extensively studied for Chinese sequence labeling.
For instance, TENER \cite{yan2019tener} adopts Transformer encoder to model character-level features for Chinese NER.
Glyce \cite{meng2019glyce} uses BERT to capture the contextual representation combined with glyph embeddings for Chinese sequence labeling.

\paragraph{Lexicon-based Methods}
In recent studies, lexicon knowledge has been applied to improve model performance.
There are two mainstream categories to the work of lexicon enhancement.
The first aims to enhance the original BERT with implicit boundary information by using the multi-granularity word masking mechanism.
BERT-wwm \cite{cui2021pre} and ERNIE \cite{sun2019ernie} are representatives of this category,
which propose to mask tokens, entities, and phrases as the mask units in the masked language modeling (MLM) task to learn the coarse-grained lexicon information during pre-training.
ERNIE-Gram \cite{xiao-etal-2021-ernie}, an extension of ERNIE, utilizes statistical boundary information for unsupervised word extraction to support masked word prediction,
The second category, which includes ZEN \cite{diao-etal-2020-zen}, EEBERT \cite{jia-etal-2020-entity}, and LEBERT \cite{liu-etal-2021-lexicon},
exploits the potential of directly injecting lexicon information into BERT via extra modules,
leading to better performance but is limited in pre-defined external knowledge.
Our work follows the first line of work, most similar to ERNIE-Gram.
However, different from ERNIE-Gram, we do not discretize the real-valued statistical information extracted from corpus, but adopt a regression manner to leverage the information fully.


\section{Method}
Figure \ref{fig:model} shows the overall architecture of our unsupervised boundary-aware pre-trained language model,
which mainly consists of three components:
1) boundary information extractor for unsupervised statistical boundary information mining,
2) boundary-aware representation to integrate statistical information at the character-level,
and 3) boundary-aware BERT learning which injects boundary knowledge into the internal layer of BERT.
In this section, we first focus on the details of the above components,
and then introduce the fine-tuning method for Chinese sequence labeling.

\subsection{Boundary Information Extractor}
\label{sec:unsper-bound-ext}
Statistical boundary information has been shown with a positive influence on a variety of Chinese NLP tasks \cite{Song12usinga, higashiyama-etal-2019-incorporating, ding-etal-2020-coupling, xiao-etal-2021-ernie}.
We follow this line of work, designing a boundary information extractor to mine statistical information from a large raw corpus in an unsupervised way.

The overall flow of the extractor includes two steps:
I) First, we collect all N-grams from the raw corpus to build a dictionary $\mathcal{N}$,
in which we count the frequencies of each N-gram and filter out the low frequencies items; 
II) second, considering that word frequency is insufficient for representing the flexible boundary relation in the Chinese context,
we further compute two unsupervised indicators which can capture most of the boundary information in the corpus.
In the following, we will describe these two indicators in detail.

\paragraph{Pointwise Mutual Information (PMI)}
Given an N-gram, we split it into two sub-strings and compute the mutual information (MI) between them as a candidate.
Then, we enumerate all sub-string pairs and choose the minimum MI as the overall PMI to estimate the tightness of the N-gram.
Let $g = \{c_1 ... c_m\}$ be an N-gram that consists of $m$ characters, we calculate PMI using this formula:
\begin{equation}
  \label{eq:PMI}
  \small
  \texttt{PMI}(g) = \min_{i\in[1:m-1]} \{ \frac{p(g)}{p(c_1...c_i) \cdot p(c_{i+1}...c_m)} \},
\end{equation}
where $p(\cdot)$ denotes the probability over the corpus.
Note that, when $m = 1$, the corresponding PMI is constantly equal to 1.
The higher PMI indicates that the N-gram (e.g., "贝克汉姆 (Beckham)")
has a similar occurrence probability to the sub-string pair (e.g., "贝克 (Beck)" and "汉姆 (Ham)"),
leading to a higher association between internal sub-string pairs,
which makes the N-gram more likely to be a word/entity.
In contrast, a lower PMI means the N-gram (e.g., "克汉(Kehan)") is possibly an invalid word/entity.



\paragraph{Left and Right Entropy (LRE)}
Given an N-gram $g$, we first collect a left-adjacent character set $S_m^l=\{c_1^l, ..., c_{n_l}^l \}$ with $n_l$ characters.
Then, we utilize the conditional probability between $g$ and its left adjacent characters in $S_m^l$ to compute the left entropy (LE),
which measures sufficient boundary information.
LE can be defined as:
\begin{equation}
  \label{eq:le}
  \texttt{LE}(g) = -{\sum}_{i}^{n_l} p(c^l_i g | g) \log p(c^l_i g | g).
\end{equation}
Similar to LE, we further collect a right adjacent set $S_m^{r} = \{c^r_1,...,c^r_{n_r}  \}$ with $n_r$ characters to calculate the right entropy (RE) for the N-gram $g$:
\begin{equation}
  \label{eq:re}
  \texttt{RE}(g) = -{\sum}_{i}^{n_r} p(g c^r_i | g) \log p(g c^r_i | g).
\end{equation}

Intuitively, LRE represents the abundance of neighboring characters for the N-gram.
With a lower LRE, the N-gram (e.g., "汉姆 ") has a more fixed context,
indicating it is more likely to be a part of a phrase or entity.
Conversely, the N-gram with a higher LRE (e.g., "贝克汉姆") will interact more with context, which prefers to be an independent word or phrase.

Finally, we utilize PMI and LRE to measure the flexible boundary relations in the Chinese context,
and then update each N-gram in $\mathcal{N}$ with the unsupervised statistical indicators above.

\subsection{Boundary-Aware Representation}
\label{sec:encode}
By using the boundary information extractor, we can obtain an N-gram dictionary $\mathcal{N}$ with unsupervised statistical boundary information.
Unfortunately, since the context independence and the high relevance to N-gram,
previous works \cite{ding-etal-2020-coupling, xiao-etal-2021-ernie} use such statistical features for word extraction only,
which ignore the potential of statistical boundary information in representation learning.
To alleviate this problem, we propose boundary-aware representation,
a highly extensible method, to fully benefit from the statistical boundary information for representation learning.

To achieve boundary-aware representation, we first build contextual N-gram sets from the sentence.
As shown in Figure \ref{fig:model} (b), given a sentence $x = \{c_1, c_2, ..., c_n\}$ with $n$ characters and the maximum N-gram length $N$,
we extract all N-grams that include $c_i$ as the contextual N-gram set $S^{c}_i = \{\bm{c_i}, \bm{c_i}c_{i+1}, \cdots, c_{i-N+1} ... \bm{c_i}\}$ for character $c_i$.
Then, we design a composition method to integrate the statistical features of N-grams in $S^{c}_i$ by using specific conditions and rules,
aiming to avoid the sparsity and contextual independence limitations of statistical information.

Concretely, we divide the information composition method into PMI and entropy representation.
First, we concatenate the PMI of all N-grams in $S^{c}_i$ to generate PMI representation:
\begin{equation}
  \small
  \begin{split}
    \mathbf{e}_i^{p} = & \texttt{PMI}(\mybox{$c_i$}) \\
    \oplus & \texttt{PMI}(\mybox{$c_i$}c_{i+1}) \oplus \texttt{PMI}(c_{i-1}\mybox{$c_i$}) \\
    \oplus & \texttt{PMI}(\mybox{$c_i$}c_{i+1}c_{i+2}) \oplus \cdots \oplus \texttt{PMI}(c_{i-2}c_{i-1}\mybox{$c_i$})\\
    & ~~~~~~~~~~~~~~~~~~~~~~~~~~~ \cdots \cdots  \\
    \oplus & \texttt{PMI}(\mybox{$c_i$} ... c_{i+N-1}) \oplus \cdots \oplus \texttt{PMI}(c_{i-N+1} ... \mybox{$c_i$}),
  \end{split}
\end{equation}
where $\mathbf{e}_i^{p} \in \mathbb{R}^{a}$, and $a = 1+2+\cdots+N$ is the number of the N-grams that contain $c_i$.
Note that the position of each N-gram is fixed in PMI representation.
We strictly follow the order of N-gram length and the position of $c_i$ in N-gram to concatenate their corresponding PMI,
ensuring that the position and context information can be encoded into $\mathbf{e}_i^{p}$.

Entropy representation focuses on the contextual interactions of each character.
When $c_i$ is the border of N-grams in $S^{c}_i$,
we separately aggregate the LE and RE as left and right entropy representation:
\begin{equation}
  \small
  \begin{split}
    \mathbf{e}_i^{le} =& \texttt{LE}(\mybox{$c_i$}) \oplus \texttt{LE}(\mybox{$c_i$}c_{i+1}) \oplus \cdots \oplus \texttt{LE}(\mybox{$c_i$}...c_{i+N-1}),\\
    \mathbf{e}_i^{re} =& \texttt{RE}(\mybox{$c_i$}) \oplus \texttt{RE}(c_{i-1}\mybox{$c_i$}) \oplus \cdots \oplus \texttt{RE}(c_{i-N+1}...\mybox{$c_i$}),
  \end{split}
\end{equation}
where $\mathbf{e}_i^{le} \in \mathbb{R}^{b}$, $\mathbf{e}_i^{re} \in \mathbb{R}^{b}$, and $b = N$\footnote{We only consider the N-grams where $c_i$ is on the boundary, which means that for each direction the number of N-grams is equal to the maximum N-gram length.}
is the number of integrated N-grams.
Similar to PMI representation, the position of each N-gram in $\mathbf{e}_i^{le}$ and $\mathbf{e}_i^{le}$ is fixed and symmetric.
Therefore, the boundary-aware representation $\mathbf{e}_i$ of $c_i$ can be formalized as:
\begin{equation}
  \label{eq:boundary}
  \mathbf{e}_i = \mathbf{e}_i^{le} \oplus \mathbf{e}_i^{p} \oplus \mathbf{e}_i^{re},
\end{equation}
where $\mathbf{e}_i \in \mathbb{R}^{a + 2b}$.
Finally, by composing multi-granularity statistical boundary information in a specific order,
we are able to obtain the boundary-aware representation,
which explicitly contains the boundary and context information.

\begin{figure}
  \centering
  \includegraphics[scale=0.47]{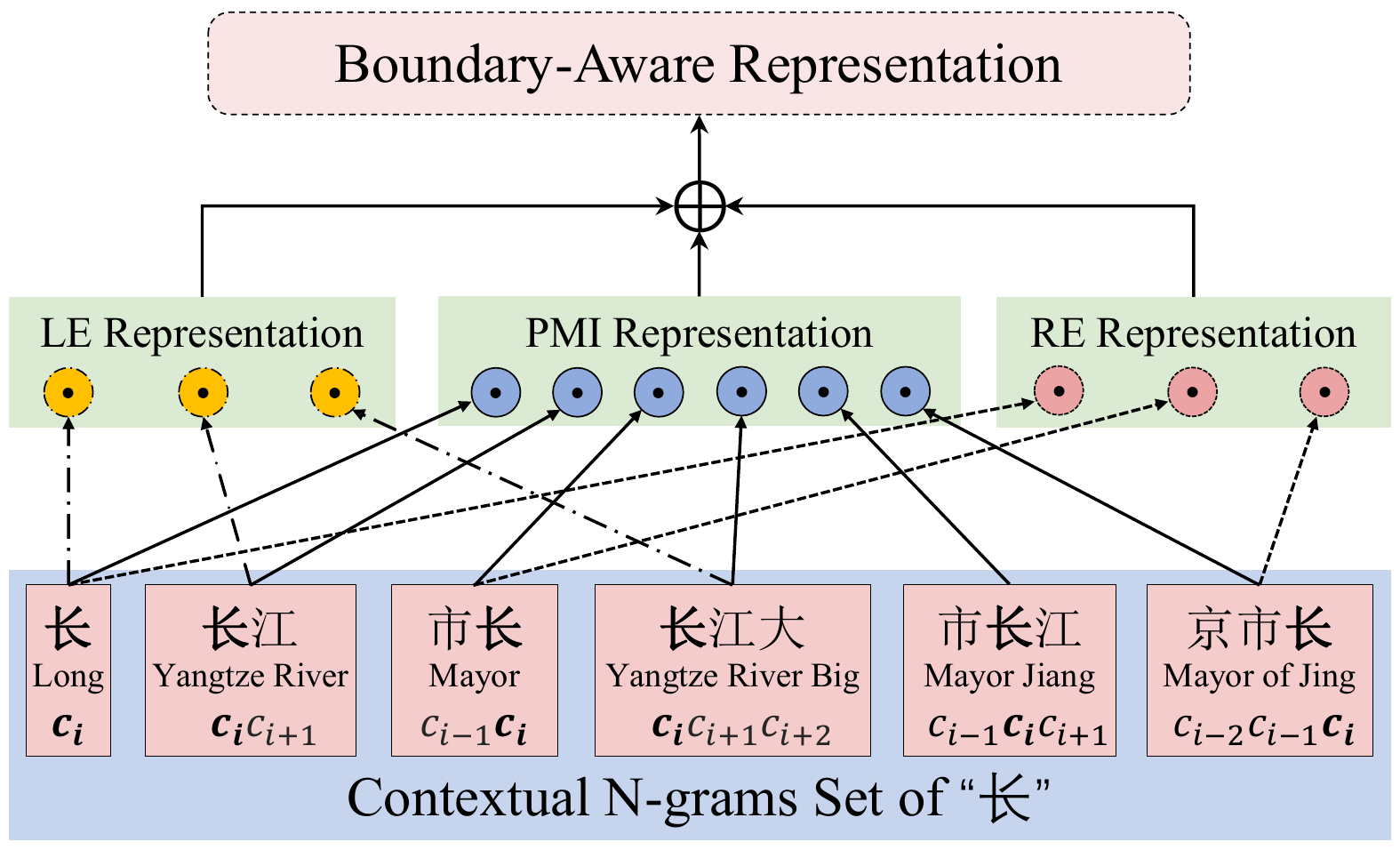}
  \caption{The illustration of boundary-aware representation for character "长" in text "南京市长江大桥".}
  \label{fig:boundary}
\end{figure}

Figure \ref{fig:boundary} shows an example of the boundary-aware representation.
Given a sentence "南京市长江大桥 (Nanjing Yangtze River Bridge)" and a maximum N-gram length $N = 3$,
we first build a contextual N-gram set for the character "长 (Long)".
Then, we integrate the PMI of all N-grams in a specific order (from N-gram "长" to "京市长 (Mayor of Jing)") to compute PMI representation. 
Furthermore, left and right entropy representations are also calculated in a particular order (from N-gram "长" to "长江大  (Yangtze River Big)" and "京市长", respectively).
Finally, we concatenate the above features to produce the overall boundary-aware representation of the character "长".


\subsection{Boundary-Aware BERT Learning}
\label{sec:bound-learning}
Boundary-aware BERT is a variant of BERT, enhanced with boundary information simply and effectively.
In this subsection, we describe how the boundary information can be integrated into BERT during pre-training by boundary-aware learning.


\paragraph{Boundary-Aware Objective}

As mentioned in Section \ref{sec:encode}, given a sentence $x$ with character-length $n$, we can compute the corresponding boundary-aware representation $\mathbf{E} = \{\mathbf{e_1}, ..., \mathbf{e_n}  \}$.
Then, we transfer the BERT feature into the boundary information space and approximate it to $\mathbf{E}$ for boundary-aware learning.
Moreover, \citet{liu-etal-2021-lexicon} shows that encoding basic lexical knowledge in the shallow BERT layers is a more effective approach.
Hence, we use the hidden features $\mathbf{H}^l = \{\mathbf{h}^l_1, ..., \mathbf{h}^l_n\}$ of the $l$-th shallow layer to achieve the boundary-aware objective:
\begin{equation}
  \label{eq:BA}
  \mathcal{L}_\texttt{BA} = {\sum}_{i}^n \texttt{MSE}(\mathbf{W}_{B} \mathbf{h}^l_i, \mathbf{e_i}),
\end{equation}
where $\mathtt{MSE}(\cdot)$ denotes the mean square error loss.
$\mathbf{W}_{B}$ is a trainable matrix used to project BERT representation into boundary information space.

Previous classification-based word-level masking methods use statistical information as thresholds to filter valid words for masked word prediction.
Unlike the above works, we softly utilize such information in a regression manner,
avoiding possible errors in empirically filtering valid tags,
thereby fully exploring the potential of this information.

\paragraph{Pre-training}
Following \citet{jia-etal-2020-entity} and \citet{gao-callan-2021-condenser},
we opt to initialize our model with a pre-trained BERT model released by Google\footnote{https://github.com/google-research/bert}
and randomly initialize the other parameters,
alleviating the enormous cost of training BABERT from scratch.
In particular, we discard the next sentence prediction task during pre-training, which is confirmed to be not essential for the pre-trained language models \cite{lan2019albert, liu2019roberta}.
The total pre-training loss of BABERT can be formalized as:
\begin{equation}
  \mathcal{L}_\texttt{pre} = \mathcal{L}_\texttt{MLM} + \mathcal{L}_\texttt{BA},
\end{equation}
where $\mathcal{L}_\texttt{MLM}$ is the standard objective of MLM task.

\subsection{Fine-tuning for Sequence Labeling}
\paragraph{Straightforward Fine-tuning}
As shown in Figure \ref{fig:model} (c), because BABERT has the same architecture as BERT,
we can adopt the identical procedure that BERT uses for fine-tuning, where the output of BABERT can be used as the contextual character representation for sequence labeling.
Concretely, given a sequence labeling dataset $\mathcal{D} = \{ (x_j, y_j) \}_{j=1}^N$,
where $y_j$ is the label sequence of $x_j$,
we utilize the output of BABERT and a CRF layer to calculate the sentence-level output probability $p(y_j|x_j)$,
which is exactly the same as \citet{liu-etal-2021-lexicon}.
The negative log-likelihood loss for training can be defined as:
\begin{equation}
  \label{eq:finetune}
  \mathcal{L}_\texttt{sq} = - {\sum}_j^N \log p(y_j| x_j),
\end{equation}
At the inference stage, we use the Viterbi algorithm \cite{viterbi1967error} to generate the final label sequence. 

\paragraph{Combining with Supervised Lexicon Features}
We can naturally combine BABERT with other supervised lexicon-based methods because of the unsupervised setting of BABERT.
To this end, we propose a lexicon-enhanced BABERT (\textbf{BABERT-LE}) for the fine-tuning stage,
which utilizes the lexicon adapter proposed by \citet{liu-etal-2021-lexicon} to incorporate external lexicon knowledge into BABERT feature:
\begin{equation}
  \mathbf{\hat{h}}_i = \texttt{LA}(\mathbf{h}_i, S_i^{lex}),
\end{equation}
where $\texttt{LA}(\cdot)$ is the lexicon adapter,
$S_i^{lex}$ is a set of related N-gram embeddings of character $c_i$,
and $\mathbf{\hat{h}}_i$ is the lexicon-enhanced version of original BERT feature $\mathbf{h}_i$.
We apply the lexicon adapter after the $l$-th layer to be consistent with boundary-aware learning.
Finally, BABERT-LE performs a similar fine-tuning procedure as BABERT for training:
\begin{equation}
  \label{eq:le_finetune}
  \small
  \mathcal{L}_\texttt{lex} = - {\sum}_j^N \log p(y_j| x_j, [S_1^{lex},...,S_{n_j}^{lex}]).
\end{equation}

\section{Experiments}

\subsection{Datasets}
Following previous works \cite{devlin2018bert, xiao-etal-2021-ernie},
we draw the mixed corpus of Chinese Wikipedia\footnote{https://zh.wikipedia.org/wiki/} and Baidu Baike\footnote{https://baike.baidu.com/} as our pre-training corpus, which contains 3B tokens and 62M sentences. 
To further confirm the effectiveness of our proposed method for Chinese sequence labeling,
we evaluate BABERT on ten benchmark datasets of three representative tasks:

\paragraph{Chinese Word Segmentation}
We use three CWS benchmarks to evaluate our BABERT.
Penn Chinese TreeBank version 6.0 (CTB6) is from \citet{xue2005penn},
and MSRA and PKU are from SIGHAN 2005 Bakeoff \cite{emerson-2005-second}.

\paragraph{Part-Of-Speech Tagging}
For Chinese POS tagging, we conduct experiments on CTB6 \cite{xue2005penn} and the Chinese part of Universal Dependencies (UD) \cite{nivre2016universal}.
The UD dataset uses two different POS tagsets, which are universal and language-specific tagsets.
We follow \citet{shao-etal-2017-character},
referring to the corpus with the two tagsets as UD1 and UD2, respectively.

\paragraph{Named Entity Recognition}
For the Chinese NER task, we conduct experiments on OntoNotes 4.0 (Onto4) \cite{weischedel2011ontonotes} and News datasets \cite{jia-etal-2020-entity}, both of which are from the standard newswire domain.
Moreover, we evaluate BABERT in the internet novel (Book) and financial report (Finance) domains \cite{jia-etal-2020-entity} to further verify the robustness of our method.

\begin{table}[]
  \centering
  {
  \begin{tabular}{@{}ll|lll@{}}
  \toprule
                       &         & Train   & Dev   & Test   \\ \midrule
  \multirow{3}{*}{CWS} & CTB6    & 23401   & 2078  & 2795   \\
                       & MSRA    & 86924   & ~~~-  & 3985   \\
                       & PKU     & 19056   & ~~~-  & 1944   \\ \hline
  \multirow{2}{*}{POS} & CTB6    & 23401   & 2078  & 2795   \\
                       & UD1/2   & 3997    & 500   & 500    \\ \hline
  \multirow{4}{*}{NER} & Onto4    & 15724   & 4303  & 4346   \\
                       & Book    & 6675    & 2551  & 2551   \\
                       & News    & 5179    & 610   & 708    \\
                       & Finance & 46364   & ~~~-  & 2000   \\ \bottomrule
  \end{tabular}}
  \caption{
  The statistics of sentence number for the benchmark datasets.
  For the datasets without development sections, we randomly select 10\% sentences from the corresponding training set as the development set.
  }
  \label{tab:datasets}
\end{table}

The statistics of the benchmark datasets are shown in Table \ref{tab:datasets}.
For a fair comparison, we split these datasets into training,
development, and test sections following previous works \cite{jia-etal-2020-entity, liu-etal-2021-lexicon}.
Note that MSRA, PKU, and Finance do not have development sections.
Therefore, we randomly select 10\% instances from the training set as the development set for these datasets.

\setlength{\arrayrulewidth}{0.5pt}
\begin{table*}[]
  \centering
  \small
  \begin{tabu}{@{}l|ccc|ccc|cccc|c@{}}
    \toprule
    \multirow{2}{*}{} & \multicolumn{3}{c|}{CWS} & \multicolumn{3}{c|}{POS} & \multicolumn{4}{c|}{NER} & \multirow{2}{*}{Avg}  \\
                            & CTB6       & MSRA       & PKU        & CTB6       & UD1        & UD2        & Onto4      & Book       & News       & Finance    &            \\ \hline \hline
    \multicolumn{12}{c}{(I) Pre-Trained Language Model} \\ \hline \hline
    BERT                    & 97.35      & 98.22      & 96.26      & 94.72      & 95.04      & 94.89      & 80.98      & 76.11      & 79.15      & 85.31      & 89.80      \\ \tabucline[0.4pt on 4pt off 4pt]
   ~BABERT                  & 97.45      & 98.44      & \bf{96.70} & \bf{95.05} & \bf{95.65} & \bf{95.54} & \bf{81.90} & 76.84      & 80.27      & \bf{86.89} & \bf{90.47} \\ \hline
    BERT-wwm                & 97.39      & 98.31      & 96.51      & 94.84      & 95.50      & 95.41      & 80.87      & 76.21      & 79.26      & 84.97      & 89.93      \\
    ERNIE                   & 97.37      & 98.25      & 96.30      & 94.90      & 95.28      & 95.12      & 80.38      & 76.56      & \bf{80.36} & 86.03      & 90.06      \\
    ERNIE-Gram              & 97.28      & 98.27      & 96.36      & 94.93      & 95.26      & 95.16      & 80.96      & \bf{77.19} & 79.96      & 85.31      & 90.07      \\
    ZEN$^{\ddag}$           & 97.33      & 98.28      & 96.55      & 94.76      & 95.55      & 95.54      & 80.06      & 75.67      & 80.17      & 85.05      & 89.90      \\ \tabucline[0.4pt on 4pt off 4pt]
   ~NEZHA$^\dag$            & \bf{97.53} & \bf{98.61} & 96.67      & 94.98      & 95.57      & 95.52      & 81.74      & 77.03      & 79.81      & 85.15      & 90.26 \\ \hline \hline
    \multicolumn{12}{c}{(II) Supervised Lexicon Enhanced Model} \\ \hline \hline
 BERT-LE$^{\ddag\spadesuit}$& 97.44      & 98.41      & 96.70      & 94.92      & 95.49      & 95.42      & 81.59      & 77.05      & 80.29      & 86.47      & 90.38      \\ \tabucline[0.4pt on 4pt off 4pt]
~BABERT-LE$^\ddag$          & \bf{97.56} & \bf{98.63} & \bf{96.84} & \bf{95.24} & \bf{95.74} & \bf{95.70} & \bf{82.35} & \bf{78.36} & \bf{80.86} & \bf{87.25} & \bf{90.85} \\
    \multicolumn{1}{r|}{$(\Delta_{\text{BABERT}})$~}  & +0.11      & +0.19      & +0.14      & +0.21      & +0.09      & +0.16      & +0.45      & +1.52      & +0.59     & +0.36      & +0.38      \\ \hline
    Lattice$^\ddag$         & 96.10      & 97.80      & 95.80      & -          & -          & -          & 73.88      & -          & -          & -          & -          \\
    MEM-BERT$^\ddag$        & 97.16      & 98.28      & 96.51      & 94.82      & 95.60      & 95.46      & -          & -          & -          & -          & -          \\
    MEM-ZEN$^\ddag$         & 97.25      & 98.40      & 96.53      & 94.87      & 95.69      & 95.49      & -          & -          & -          & -          & -          \\
    EEBERT$^\ddag$          & -          & -          & -          & -          & -          & -          & -          & 77.58      & 76.66      & 87.05      & -          \\ \bottomrule
  \end{tabu}
  \caption{
    The overall results on three Chinese sequence labeling tasks, where we report the F1-score on the test set.
    $\ddag$ denotes external knowledge is used. $\dag$ denotes that large-scale pre-training corpus is used.
    $\spadesuit$ indicates that we reproduce LEBERT in a similar way for fair comparisons.
  }
  \label{tab:main}
\end{table*}

\subsection{Experimental Settings}

\paragraph{Hyperparameters}
During pre-training, we use the hyperparameters of BERT$_{\rm BASE}$ to initialize BABERT and Adam \cite{kingma2014adam} for optimizing.
The number of BERT layers $L$ is $12$, with $12$ self-attention heads,
768 dimensions for hidden states, and 64 dimensions for each head.
The batch size is set to $32$, the learning rate is $1e$-$4$ with a warmup ratio of $0.1$, and the max length of the input sequence is 512.
To extract unsupervised boundary information, we set the maximum N-gram length $N$ to $4$
\footnote{According to our preliminary experiments, when $N$ is less than 4, the performance will be significant decrease. When $N$ is set to 5, the improvement is limited, while the time complexity will increase. Thus, we set $N$ to 4.}
and the frequency filtering threshold to 50.
Then we use the $3$-th BERT layer to compute boundary-aware objective.
BABERT has no extra modules, which is why the parameter size and model architecture are the same as those of BERT$_{\rm BASE}$.
Finally, we train the BABERT on 8 NVIDIA Tesla V100 GPUs with 32GB memory.

For Chinese sequence labeling, we empirically set hyperparameters based on previous studies \cite{jia-etal-2020-entity,liu-etal-2021-lexicon} and preliminary experiments.
The batch size is 32, the max sequence length is 256, and the learning rate is fixed to $2e$-$5$.

\paragraph{Baselines}
To verify the effectiveness of our proposed BABERT, we build systems on the following methods to conduct fair comparisons:

\begin{itemize}
  \item \textbf{BERT} is the Chinese version BERT$_{\rm BASE}$ model released by Google.

  \item \textbf{BERT-wwm} performs segmentation on the corpus and further conduct word-level masking in pre-training \cite{cui2021pre}.

  \item \textbf{ERNIE} is an extension of BERT, which leverages external lexicons for word-level masking \cite{sun2019ernie}.

  \item \textbf{ERNIE-Gram} is an extension of ERNIE,
  which alleviates the limitations of external lexicons by using statistical information for entity and phrase extraction \cite{xiao-etal-2021-ernie}.

  \item \textbf{ZEN} uses an extra N-gram encoder to integrate external lexicon knowledge into BERT during pre-training \cite{diao-etal-2020-zen}.


  \item \textbf{NEZHA} leverages functional relative positional encoding, supervised word-level masking strategy, and enormous training data\footnote{NEZHA uses three large corpora, including Chinese Wikipedia, Baidu Baike, and Chinese News, which contain 11B tokens and are four times more than us.}
  to enhance vanilla BERT \cite{wei2019nezha}.

  \item \textbf{BERT-LE} is a lexicon-enhanced BERT \cite{liu-etal-2021-lexicon},
  which introduces a lexicon adapter between BERT layers to incorporate external lexicon embeddings.
  We strictly follow.\citet{liu-etal-2021-lexicon} to reimplement it with open-source word embeddings\footnote{https://ai.tencent.com/ailab/nlp/en/embedding.html}
\end{itemize}

\subsection{Main Results}
The overall Chinese sequence labeling results are shown in Table \ref{tab:main}.
We report the F1-score of the test datasets on CWS, POS, and NER tasks.
Here, we first compare our BABERT with various Chinese pre-trained language models to evaluate its effectiveness.
Then, we compare BABERT-LE with other supervised lexicon-based methods to show the potential of BABERT in combining with external lexicon knowledge.

First, we examine the F1 values of the BERT baseline.
As shown, BERT obtains comparable results on all Chinese sequence labeling tasks,
which is similar to that of \citet{diao-etal-2020-zen}, \citet{tian-etal-2020-joint-chinese} and \citet{liu-etal-2021-lexicon}.
BABERT significantly outperforms BERT, resulting in an increase of $90.47 - 89.80 = 0.67$ on average.
This observation clearly indicates the advantage of introducing boundary information into BERT pre-training.

Compared with various BERT extensions,
our BABERT can achieve competitive performances as a whole.
First, in comparison with BERT-wwm, ERNIE, ERNIE-gram, and ZEN,
which leverage external lexicons that include high-frequency words for pre-training,
BABERT outperforms all of them by averaging $\frac{0.54+0.41+0.40+0.57}{4} = 0.48 $  point, and achieves top scores on eight of the ten benchmarks.
This result is consistent with our intuition that directly exploiting a supervised lexicon can only achieve good performance in specific tasks,
indicating the limitation of these methods when the chosen lexicon is incompatible with the target tasks.
Second, we find that BABERT surpasses NEZHA in the average F1 values,
indicating that the boundary information is more critical than the data scale for Chinese sequence labeling.

Then, we compare our method with supervised lexicon-based methods.
Lattice-based methods \cite{zhang-yang-2018-chinese, yang-etal-2019-subword} are the first to integrate word features into neural networks for Chinese sequence labeling.
MEM \cite{tian-etal-2020-joint-chinese, tian-etal-2020-improving-chinese}
designs an external memory network after the BERT encoder to incorporate lexicon knowledge.
EEBERT \cite{jia-etal-2020-entity} builds entity embeddings from the corpus and further utilizes them in the multi-head attention mechanism.
The results are shown in Table \ref{tab:main} (II).
All the above methods lead to significant improvements over the base BERT model,
which shows the effectiveness of external lexicon knowledge.
Moreover, BABERT can achieve comparable performance with the above methods,
which further demonstrates the potential of our unsupervised manner. 

BABERT learns boundary information from unsupervised statistical features with vanilla BERT,
which means it has excellent scalability to fuse with other BERT-based supervised lexicon models.
As shown, we can see that our BABERT-LE achieves further improvements and state-of-the-art performances on all tasks,
showing the advantages of our unsupervised setting and boundary-aware learning.
Interestingly, compared with MEM-ZEN, BABERT-LE has larger improvements over their corresponding baselines.
One reason might be that both ZEN and the memory network module exploits supervised lexicons,
which leads to a duplication of introduced knowledge.



\subsection{Analysis}
\label{sec:analysis}
In this subsection, we conduct detailed experimental analyses for an in-depth comprehensive understanding of our method.


\begin{table}[t]
  \centering
  \resizebox{0.48\textwidth}{!}{%
  \begin{tabu}{@{}l|ccc|ccc@{}}
  \toprule
             & \multicolumn{3}{c|}{CWS-PKU}                    & \multicolumn{3}{c}{NER-Onto4}                     \\
             & 10             & 50             & 100            & 10             & 50             & 100            \\ \midrule
  BERT       & 83.97          & 87.93          & 88.25          & 14.87          & 42.37          & 57.95          \\ \tabucline[0.4pt on 4pt off 4pt]
  ~BABERT     & \textbf{84.73} & \textbf{89.45} & \textbf{90.02} & \bf{32.07}     & \bf{46.55}     & \textbf{60.58} \\ \hline
  BERT-wwm   & 84.70          & 88.05          & 88.84          & 12.75          & 43.09          & 59.44          \\
  ERNIE      & 84.31          & 87.04          & 88.20          & 19.86          & 42.96          & 50.81          \\
  ERNIE-Gram & 84.01          & 86.62          & 87.99          & 28.39          & 45.88          & 60.01          \\ 
  NEZHA      & 84.40          & 88.70          & 89.73          & 14.45          & 44.10          & 59.20          \\ \bottomrule
  \end{tabu}
  }
  \caption{Few-shot results on PKU and Onto4, using 10, 50, and 100 instances of the training data.}
  \label{tab:few-shot}
\end{table}

\paragraph{Few-Shot Setting}
To further verify the effectiveness of BABERT, we conduct experiments under the few-shot setting,
where we randomly sample 10, 50, and 100 instances of the original training data from PKU (CWS) and Onto4 (NER).
For fair comparisons, we compare BABERT with the pre-trained language models without external supervised knowledge.
The results are presented in Table \ref{tab:few-shot}.
As the size of training data is reduced, the performance drops significantly,
indicating that the performances of such models rely on the scale of the labeled training data.
Nevertheless, BABERT achieves top scores under each setting and significantly outperforms vanilla BERT,
demonstrating the potential of injecting unsupervised boundary information by using regression-based boundary-aware learning,
which effectively alleviates the low-resource problem.

\begin{table}[t]
  \centering
  \begin{tabular}{@{}r|cccc@{}}
  \toprule
  \multicolumn{1}{l|}{}       & Onto4       & Book        & News        & Finance \\ \midrule
  \multicolumn{1}{l|}{BABERT} & \bf{81.90}  & \bf{76.84}  & \bf{80.27}  & \bf{86.89}   \\ \midrule
  + T-test                    & 81.77       & 76.65       & 80.12       & 86.23   \\
  - PMI                       & 81.12       & 76.28       & 79.51       & 85.56   \\
  - LRE                       & 81.51       & 76.47       & 79.67       & 85.94  \\ \bottomrule
  - LE                       & 81.58       & 76.62       & 79.88       & 85.94  \\ 
  - RE                       & 81.53       & 76.59       & 79.75       & 85.90  \\ \bottomrule
  \end{tabular}
  \caption{The results of different feature combining settings on the four NER benchmark datasets.}
  \label{tab:ana-feature}
\end{table}

\begin{table}[t]
  \centering
  \begin{tabular}{@{}r|c|c|cc@{}}
  \toprule
  \multirow{2}{*}{Layer}   & CWS        & POS        & \multicolumn{2}{c}{NER} \\
                           & PKU        & CTB6       & Onto4      & News       \\ \midrule
  12                       & 96.38      & 94.68      & 81.02      & 79.19     \\
  6                        & 96.53      & 94.84      & 81.45      & 79.91      \\
  3                        &\bf{96.70}  & \bf{95.03} & \bf{81.90} & \bf{80.27}      \\
  1                        & 96.62      & 94.90      & 81.59      & 79.52      \\ \bottomrule
  \end{tabular}
  \caption{The influence of boundary information learn at different layers of BERT model.}
  \label{tab:layer}
\end{table}

\paragraph{Choice of Statistical Features}
As mentioned in Section \ref{sec:encode}, we use PMI and LRE to model unsupervised boundary-aware representation.
In addition to these features, T-test \cite{xiao-etal-2021-ernie} is another popular choice that can be utilized.
Thus, we conduct ablation experiments to check the effectiveness of these features.
We analyze five feature combining settings, including PMI (-LRE) and LRE (-PMI) alone, the ablation study of LRE (-LE and -RE), 
and our two indicators with additional T-test features being concatenated at the end of the original boundary-aware representation (+T-test).
As shown in Table \ref{tab:ana-feature}, the combination of PMI and LRE can achieve the best results,
discarding either of them will result in decreased performance.
Besides, right entropy is more important than left entropy according to our results,
which may be that the right entropy is more compatible with the reading characteristics of Chinese.
Interestingly, adding T-test does not bring further improvements.
One possible reason is that the T-test is essentially similar to the entropy measure of 2-grams,
which has already been injected into our BABERT model.

\paragraph{Boundary Information Encoding Layer}
Previous works \cite{jawahar2019does, liu-etal-2021-lexicon} exploit the fact that different BERT layers would generate different concept representations.
The shallow BERT layers are more likely to capture basic lexicon information,
while the top layers focus on the semantic representation.
We empirically set $l$ in $\{1, 3, 6, 12\}$ to explore the effect of computing boundary-aware loss by the hidden features $\mathbf{H}^l$ of different BERT layers on Chinese sequence labeling tasks.
Table \ref{tab:layer} shows the results.
We can see that the best F1-score can be achieved when $l=3$ on all datasets, 
 which indicates that the BABERT still needs sufficient parameters to learn the basic boundary information.
Interestingly, the BABERT performs poorly when $l=12$,
which might be due to a conflict between the MLM loss and our boundary-aware regression loss during pretraining.

\paragraph{Qualitative Analysis}
To explore how BABERT improves the performance for Chinese sequence labeling, we conduct qualitative analysis on the News test dataset, which consists of four different subdomains, namely game (GAM), entertainment (ENT), lottery (LOT) and finance (FIN).
The results are shown in Table \ref{tab:qualitative}.
We can see that compared with other pre-trained language models, BABERT can obtain consistent improvement in all domains with unsupervised statistical boundary information,
while the other models only improve performance on specific domains.
Moreover, as shown in Table \ref{tab:case-study}, we also give an example from the game domain to further demonstrate the effectiveness of our method.
BABERT is the only model that correctly recognizes all entities.
In particular, the prediction of BABERT for the entity "WCG2011$_\texttt{org}$" indicates the potential of boundary information.

\begin{table}[t]
  \centering
  \small
  \begin{tabu}{@{}l|cccc@{}}
  \toprule
                              & ENT       & GAM        & LOT        & FIN  \\ \midrule
  BERT                        & 89.50     & 74.35      & 83.62      & 79.78 \\\tabucline[0.4pt on 4pt off 4pt]
  ~BABERT                      & \bf{89.86}     & \bf{75.65}      & 83.74      & 80.22 \\\hline
  BERT-wwm                    & 83.82     & 75.39      & 81.18      & \bf{81.38} \\
  ERNIE                       & 89.75     & 74.76      & 86.69      & 80.20 \\
  ERNIE-Gram                  & 88.95     & 74.58     & 84.21      & 79.83 \\
  NEZHA                       & 89.47     & 72.78      & \bf{87.12}      & 79.75 \\ \bottomrule
  \end{tabu}
  \caption{The results of different feature combining settings on the four NER benchmark datasets.}
  \label{tab:qualitative}
\end{table}

\begin{table}[]
\centering
\small
\resizebox{\columnwidth}{!}{%
\begin{tabular}{@{}l|l@{}}
\toprule
\multirow{2}{*}{Sentence} & 如果\mybox{WCG2011}$_{\texttt{org}}$有\mybox{WAR3}$_{\texttt{game}}$和\mybox{星际2}$_{\texttt{game}}$两个项目\\
 & If WCG2021 has WAR3 and StarCraft2 these two projects \\ \midrule
BERT &  如果\wrongbox{WCG}$_{\texttt{org}}$2011有\mybox{WAR3}$_{\texttt{game}}$和\mybox{星际2}$_{\texttt{game}}$两个项目\\ \hline
BABERT &  如果\mybox{WCG2011}$_{\texttt{org}}$有\mybox{WAR3}$_{\texttt{game}}$和\mybox{星际2}$_{\texttt{game}}$两个项目\\ \hline
BERT-wwm &  如果\wrongbox{WCG2}$_{\texttt{org}}$011有\mybox{WAR3}$_{\texttt{game}}$和\mybox{星际2}$_{\texttt{game}}$两个项目\\ \hline
ERNIE &  如果\wrongbox{WCG20}$_{\texttt{org}}$11有\mybox{WAR3}$_{\texttt{game}}$和\mybox{星际2}$_{\texttt{game}}$两个项目\\ \hline
ERNIE-Gram &  如果\wrongbox{WCG}$_{\texttt{org}}$2011有\mybox{WAR3}$_{\texttt{game}}$和\mybox{星际2}$_{\texttt{game}}$两个项目\\ \hline
NEZHA &  如果WCG2011有\wrongbox{WAR3}$_{\texttt{org}}$和\mybox{星际2}$_{\texttt{game}}$两个项目\\ \bottomrule
\end{tabular}%
}
\caption{Example from the game domain. Red (Blue) represents correct (incorrect) entities.}
\label{tab:case-study}
\end{table}


\section{Conclusion}
\label{sec:conclusion}
In this paper, we proposed BABERT, a novel unsupervised boundary-aware pre-training model for Chinese sequence labeling.
In BABERT, given a Chinese sentence, we calculated boundary-aware representation with unsupervised statistical information to capture boundary information,
and directly injected such information into BERT weights during pre-training.
Unlike previous works, BABERT exploited an effective way to utilize boundary information in an unsupervised manner,
thereby alleviating the limitations of supervised lexicon-based approaches.
Experimental results on ten benchmark datasets of three different tasks illustrated that our method was highly effective and better than other Chinese pre-trained models. 
Moreover, the combination with supervised lexicon extensions could achieve further improvements and state-of-the-art results on most tasks.

\section*{Limitations}
BABERT suffers from three major limitations.
The first limitation is that in the boundary information extractor,
we empirically chose PMI and LRE. 
In addition to these indicators and the T-test measure we verified in the experimental analyses,
some alternatives that contain boundary information could be used to compute boundary-aware representation.
Thus, we plan to explore more unsupervised statistical features.
The second limitation is that we focused only on Chinese sequence labeling tasks in this work,
ignoring the potential of boundary information and BABERT in other Chinese NLP tasks.
The third one is that we only consider BABERT for Chinese.
For other languages which do not use spaces between words such as Japanese and Thai, 
we can also attempt to inject boundary information, and the effectiveness in these languages should be verified by experiments.
Further research is needed to evaluate our BABERT in future studies.

\section*{Acknowledgements}
This work is supported by grants from the National Key Research and Development Program of China (No. 2018YFC0832101) and the National Natural
Science Foundation of China (No. 62176180).


\bibliography{anthology}

\begin{thebibliography}{43}
\expandafter\ifx\csname natexlab\endcsname\relax\def\natexlab#1{#1}\fi

\bibitem[{Bellegarda(2004)}]{bellegarda2004statistical}
Jerome~R Bellegarda. 2004.
\newblock Statistical language model adaptation: review and perspectives.
\newblock \emph{Speech communication}, 42(1):93--108.

\bibitem[{Bouma(2009)}]{bouma2009normalized}
Gerlof Bouma. 2009.
\newblock Normalized (pointwise) mutual information in collocation extraction.
\newblock \emph{Proceedings of GSCL}, 30:31--40.

\bibitem[{Chiu and Nichols(2016)}]{TACL792}
Jason Chiu and Eric Nichols. 2016.
\newblock Named entity recognition with bidirectional lstm-cnns.
\newblock \emph{Transactions of the Association for Computational Linguistics},
  4(0):357--370.

\bibitem[{Clark et~al.(2020)Clark, Luong, Le, and Manning}]{clark2020electra}
Kevin Clark, Minh-Thang Luong, Quoc~V. Le, and Christopher~D. Manning. 2020.
\newblock Electra: Pre-training text encoders as discriminators rather than
  generators.
\newblock In \emph{International Conference on Learning Representations}.

\bibitem[{Cui et~al.(2021)Cui, Che, Liu, Qin, and Yang}]{cui2021pre}
Yiming Cui, Wanxiang Che, Ting Liu, Bing Qin, and Ziqing Yang. 2021.
\newblock Pre-training with whole word masking for chinese bert.
\newblock \emph{IEEE/ACM Transactions on Audio, Speech, and Language
  Processing}, 29:3504--3514.

\bibitem[{Devlin et~al.(2019)Devlin, Chang, Lee, and
  Toutanova}]{devlin2018bert}
Jacob Devlin, Ming-Wei Chang, Kenton Lee, and Kristina Toutanova. 2019.
\newblock \href {https://doi.org/10.18653/v1/N19-1423} {{BERT}: Pre-training of
  deep bidirectional transformers for language understanding}.
\newblock In \emph{Proceedings of 2019 Conference of the NAACL}, pages
  4171--4186.

\bibitem[{Diao et~al.(2020)Diao, Bai, Song, Zhang, and
  Wang}]{diao-etal-2020-zen}
Shizhe Diao, Jiaxin Bai, Yan Song, Tong Zhang, and Yonggang Wang. 2020.
\newblock {ZEN}: Pre-training {C}hinese text encoder enhanced by n-gram
  representations.
\newblock In \emph{Findings of the EMNLP 2020}, pages 4729--4740. Association
  for Computational Linguistics.

\bibitem[{Ding et~al.(2020)Ding, Long, Xu, Zhu, Xie, Wang, and
  Zheng}]{ding-etal-2020-coupling}
Ning Ding, Dingkun Long, Guangwei Xu, Muhua Zhu, Pengjun Xie, Xiaobin Wang, and
  Haitao Zheng. 2020.
\newblock Coupling distant annotation and adversarial training for cross-domain
  {C}hinese word segmentation.
\newblock In \emph{Proceedings of the 58th ACL}, pages 6662--6671. Association
  for Computational Linguistics.

\bibitem[{Emerson(2005)}]{emerson-2005-second}
Thomas Emerson. 2005.
\newblock The second international {C}hinese word segmentation bakeoff.
\newblock In \emph{Proceedings of the Fourth {SIGHAN} Workshop on {C}hinese
  Language Processing}.

\bibitem[{Gao and Callan(2021)}]{gao-callan-2021-condenser}
Luyu Gao and Jamie Callan. 2021.
\newblock Condenser: a pre-training architecture for dense retrieval.
\newblock In \emph{Proceedings of the 2021 Conference on Empirical Methods in
  Natural Language Processing}, pages 981--993. Association for Computational
  Linguistics.

\bibitem[{Higashiyama et~al.(2019)Higashiyama, Utiyama, Sumita, Ideuchi, Oida,
  Sakamoto, and Okada}]{higashiyama-etal-2019-incorporating}
Shohei Higashiyama, Masao Utiyama, Eiichiro Sumita, Masao Ideuchi, Yoshiaki
  Oida, Yohei Sakamoto, and Isaac Okada. 2019.
\newblock Incorporating word attention into character-based word segmentation.
\newblock In \emph{Proceedings of the 2019 Conference of the NAACL}, pages
  2699--2709. Association for Computational Linguistics.

\bibitem[{Jawahar et~al.(2019)Jawahar, Sagot, and Seddah}]{jawahar2019does}
Ganesh Jawahar, Beno{\^\i}t Sagot, and Djam{\'e} Seddah. 2019.
\newblock What does bert learn about the structure of language?
\newblock In \emph{Proceedings of the 57th ACL}.

\bibitem[{Jia et~al.(2020)Jia, Shi, Yang, and Zhang}]{jia-etal-2020-entity}
Chen Jia, Yuefeng Shi, Qinrong Yang, and Yue Zhang. 2020.
\newblock Entity enhanced {BERT} pre-training for {C}hinese {NER}.
\newblock In \emph{Proceedings of the 2020 Conference on EMNLP}, pages
  6384--6396. Association for Computational Linguistics.

\bibitem[{Jin and Chen(2008)}]{jin-chen-2008-fourth}
Guangjin Jin and Xiao Chen. 2008.
\newblock The fourth international {C}hinese language processing bakeoff:
  {C}hinese word segmentation, named entity recognition and {C}hinese {POS}
  tagging.
\newblock In \emph{Proceedings of the Sixth {SIGHAN} Workshop on {C}hinese
  Language Processing}.

\bibitem[{Kingma and Ba(2014)}]{kingma2014adam}
Diederik~P Kingma and Jimmy Ba. 2014.
\newblock Adam: A method for stochastic optimization.
\newblock In \emph{International Conference on Learning Representations}.

\bibitem[{Lample et~al.(2016)Lample, Ballesteros, Subramanian, Kawakami, and
  Dyer}]{lample-etal-2016-neural}
Guillaume Lample, Miguel Ballesteros, Sandeep Subramanian, Kazuya Kawakami, and
  Chris Dyer. 2016.
\newblock Neural architectures for named entity recognition.
\newblock In \emph{Proceedings of the 2016 Conference of NAACL}, pages
  260--270. Association for Computational Linguistics.

\bibitem[{Lan et~al.(2020)Lan, Chen, Goodman, Gimpel, Sharma, and
  Soricut}]{lan2019albert}
Zhenzhong Lan, Mingda Chen, Sebastian Goodman, Kevin Gimpel, Piyush Sharma, and
  Radu Soricut. 2020.
\newblock Albert: A lite bert for self-supervised learning of language
  representations.
\newblock In \emph{International Conference on Learning Representations}.

\bibitem[{Liang(2005)}]{liang2005semi}
Percy Liang. 2005.
\newblock \emph{Semi-supervised learning for natural language}.
\newblock Ph.D. thesis, Massachusetts Institute of Technology.

\bibitem[{Liu et~al.(2021)Liu, Fu, Zhang, and Xiao}]{liu-etal-2021-lexicon}
Wei Liu, Xiyan Fu, Yue Zhang, and Wenming Xiao. 2021.
\newblock Lexicon enhanced {C}hinese sequence labeling using {BERT} adapter.
\newblock In \emph{Proceedings of the 59th ACL and the 11th IJCNLP}, pages
  5847--5858. Association for Computational Linguistics.

\bibitem[{Liu et~al.(2019{\natexlab{a}})Liu, Xu, Xu, Song, and
  Zu}]{liu-etal-2019-encoding}
Wei Liu, Tongge Xu, Qinghua Xu, Jiayu Song, and Yueran Zu. 2019{\natexlab{a}}.
\newblock An encoding strategy based word-character {LSTM} for {C}hinese {NER}.
\newblock In \emph{Proceedings of the 2019 Conference of the NAACL}, pages
  2379--2389. Association for Computational Linguistics.

\bibitem[{Liu et~al.(2019{\natexlab{b}})Liu, Ott, Goyal, Du, Joshi, Chen, Levy,
  Lewis, Zettlemoyer, and Stoyanov}]{liu2019roberta}
Yinhan Liu, Myle Ott, Naman Goyal, Jingfei Du, Mandar Joshi, Danqi Chen, Omer
  Levy, Mike Lewis, Luke Zettlemoyer, and Veselin Stoyanov. 2019{\natexlab{b}}.
\newblock Roberta: A robustly optimized bert pretraining approach.
\newblock \emph{arXiv preprint arXiv:1907.11692}.

\bibitem[{Low et~al.(2005)Low, Ng, and Guo}]{low2005maximum}
Jin~Kiat Low, Hwee~Tou Ng, and Wenyuan Guo. 2005.
\newblock A maximum entropy approach to chinese word segmentation.
\newblock In \emph{Proceedings of the fourth SIGHAN workshop on Chinese
  language processing}.

\bibitem[{Ma and Hovy(2016)}]{ma-hovy-2016-end}
Xuezhe Ma and Eduard Hovy. 2016.
\newblock End-to-end sequence labeling via bi-directional {LSTM}-{CNN}s-{CRF}.
\newblock In \emph{Proceedings of the 54th ACL}, pages 1064--1074. Association
  for Computational Linguistics.

\bibitem[{Meng et~al.(2019)Meng, Wu, Wang, Li, Nie, Yin, Li, Han, Sun, and
  Li}]{meng2019glyce}
Yuxian Meng, Wei Wu, Fei Wang, Xiaoya Li, Ping Nie, Fan Yin, Muyu Li, Qinghong
  Han, Xiaofei Sun, and Jiwei Li. 2019.
\newblock Glyce: Glyph-vectors for chinese character representations.
\newblock \emph{Advances in Neural Information Processing Systems}, 32.

\bibitem[{Nivre et~al.(2016)Nivre, De~Marneffe, Ginter, Goldberg, Hajic,
  Manning, McDonald, Petrov, Pyysalo, Silveira et~al.}]{nivre2016universal}
Joakim Nivre, Marie-Catherine De~Marneffe, Filip Ginter, Yoav Goldberg, Jan
  Hajic, Christopher~D Manning, Ryan McDonald, Slav Petrov, Sampo Pyysalo,
  Natalia Silveira, et~al. 2016.
\newblock Universal dependencies v1: A multilingual treebank collection.
\newblock In \emph{Proceedings of the Tenth International Conference on
  Language Resources and Evaluation (LREC'16)}, pages 1659--1666.

\bibitem[{Peng et~al.(2004)Peng, Feng, and McCallum}]{peng2004chinese}
Fuchun Peng, Fangfang Feng, and Andrew McCallum. 2004.
\newblock Chinese segmentation and new word detection using conditional random
  fields.
\newblock In \emph{COLING 2004: Proceedings of the 20th International
  Conference on Computational Linguistics}, pages 562--568.

\bibitem[{Shao et~al.(2017)Shao, Hardmeier, Tiedemann, and
  Nivre}]{shao-etal-2017-character}
Yan Shao, Christian Hardmeier, J{\"o}rg Tiedemann, and Joakim Nivre. 2017.
\newblock Character-based joint segmentation and {POS} tagging for {C}hinese
  using bidirectional {RNN}-{CRF}.
\newblock In \emph{Proceedings of the 8-th IJCNLP}, pages 173--183. Asian
  Federation of Natural Language Processing.

\bibitem[{Shen et~al.(2016)Shen, Li, Choe, Chu, Kawahara, and
  Kurohashi}]{shen-etal-2016-consistent}
Mo~Shen, Wingmui Li, HyunJeong Choe, Chenhui Chu, Daisuke Kawahara, and Sadao
  Kurohashi. 2016.
\newblock Consistent word segmentation, part-of-speech tagging and dependency
  labelling annotation for {C}hinese language.
\newblock In \emph{Proceedings of {COLING} 2016, the 26th International
  Conference on Computational Linguistics: Technical Papers}, pages 298--308.
  The COLING 2016 Organizing Committee.

\bibitem[{Song and Xia(2012)}]{Song12usinga}
Yan Song and Fei Xia. 2012.
\newblock Using a goodness measurement for domain adaptation: A case study on
  chinese word segmentation.
\newblock In \emph{In Proceedings of the Eight International Conference on
  Language Resources and Evaluation (LREC’12}.

\bibitem[{Sun and Uszkoreit(2012)}]{sun-uszkoreit-2012-capturing}
Weiwei Sun and Hans Uszkoreit. 2012.
\newblock Capturing paradigmatic and syntagmatic lexical relations: Towards
  accurate {C}hinese part-of-speech tagging.
\newblock In \emph{Proceedings of the 50th ACL}, pages 242--252. Association
  for Computational Linguistics.

\bibitem[{Sun et~al.(2019)Sun, Wang, Li, Feng, Chen, Zhang, Tian, Zhu, Tian,
  and Wu}]{sun2019ernie}
Yu~Sun, Shuohuan Wang, Yukun Li, Shikun Feng, Xuyi Chen, Han Zhang, Xin Tian,
  Danxiang Zhu, Hao Tian, and Hua Wu. 2019.
\newblock Ernie: Enhanced representation through knowledge integration.
\newblock \emph{arXiv preprint arXiv:1904.09223}.

\bibitem[{Tian et~al.(2020{\natexlab{a}})Tian, Song, Ao, Xia, Quan, Zhang, and
  Wang}]{tian-etal-2020-joint-chinese}
Yuanhe Tian, Yan Song, Xiang Ao, Fei Xia, Xiaojun Quan, Tong Zhang, and
  Yonggang Wang. 2020{\natexlab{a}}.
\newblock Joint {C}hinese word segmentation and part-of-speech tagging via
  two-way attentions of auto-analyzed knowledge.
\newblock In \emph{Proceedings of the 58th ACL}, pages 8286--8296. Association
  for Computational Linguistics.

\bibitem[{Tian et~al.(2020{\natexlab{b}})Tian, Song, Xia, Zhang, and
  Wang}]{tian-etal-2020-improving-chinese}
Yuanhe Tian, Yan Song, Fei Xia, Tong Zhang, and Yonggang Wang.
  2020{\natexlab{b}}.
\newblock Improving {C}hinese word segmentation with wordhood memory networks.
\newblock In \emph{Proceedings of the 58th ACL}, pages 8274--8285. Association
  for Computational Linguistics.

\bibitem[{Viterbi(1967)}]{viterbi1967error}
Andrew Viterbi. 1967.
\newblock Error bounds for convolutional codes and an asymptotically optimum
  decoding algorithm.
\newblock \emph{IEEE transactions on Information Theory}, 13(2):260--269.

\bibitem[{Wei et~al.(2019)Wei, Ren, Li, Huang, Liao, Wang, Lin, Jiang, Chen,
  and Liu}]{wei2019nezha}
Junqiu Wei, Xiaozhe Ren, Xiaoguang Li, Wenyong Huang, Yi~Liao, Yasheng Wang,
  Jiashu Lin, Xin Jiang, Xiao Chen, and Qun Liu. 2019.
\newblock Nezha: Neural contextualized representation for chinese language
  understanding.
\newblock \emph{arXiv preprint arXiv:1909.00204}.

\bibitem[{Weischedel et~al.(2011)Weischedel, Pradhan, Ramshaw, Palmer, Xue,
  Marcus, Taylor, Greenberg, Hovy, Belvin et~al.}]{weischedel2011ontonotes}
Ralph Weischedel, Sameer Pradhan, Lance Ramshaw, Martha Palmer, Nianwen Xue,
  Mitchell Marcus, Ann Taylor, Craig Greenberg, Eduard Hovy, Robert Belvin,
  et~al. 2011.
\newblock Ontonotes release 4.0.
\newblock \emph{LDC2011T03, Philadelphia, Penn.: Linguistic Data Consortium}.

\bibitem[{Xiao et~al.(2021)Xiao, Li, Zhang, Sun, Tian, Wu, and
  Wang}]{xiao-etal-2021-ernie}
Dongling Xiao, Yu-Kun Li, Han Zhang, Yu~Sun, Hao Tian, Hua Wu, and Haifeng
  Wang. 2021.
\newblock {ERNIE}-gram: Pre-training with explicitly n-gram masked language
  modeling for natural language understanding.
\newblock In \emph{Proceedings of the 2021 Conference of NAACL}, pages
  1702--1715. Association for Computational Linguistics.

\bibitem[{Xue et~al.(2005)Xue, Xia, Chiou, and Palmer}]{xue2005penn}
Naiwen Xue, Fei Xia, Fu-Dong Chiou, and Marta Palmer. 2005.
\newblock The penn chinese treebank: Phrase structure annotation of a large
  corpus.
\newblock \emph{Natural language engineering}, 11(2):207.

\bibitem[{Yan et~al.(2019)Yan, Deng, Li, and Qiu}]{yan2019tener}
Hang Yan, Bocao Deng, Xiaonan Li, and Xipeng Qiu. 2019.
\newblock Tener: adapting transformer encoder for named entity recognition.
\newblock \emph{arXiv preprint arXiv:1911.04474}.

\bibitem[{Yang et~al.(2016)Yang, Teng, Zhang, and Zhang}]{Yang2016Combining}
Jie Yang, Zhiyang Teng, Meishan Zhang, and Yue Zhang. 2016.
\newblock Combining discrete and neural features for sequence labeling.
\newblock In \emph{The 17th International Conference on Intelligent Text
  Processing and Computational Linguistics (CICLing)}.

\bibitem[{Yang et~al.(2019)Yang, Zhang, and Liang}]{yang-etal-2019-subword}
Jie Yang, Yue Zhang, and Shuailong Liang. 2019.
\newblock Subword encoding in lattice {LSTM} for {C}hinese word segmentation.
\newblock In \emph{Proceedings of the 2019 Conference of the NAACL}, pages
  2720--2725. Association for Computational Linguistics.

\bibitem[{Zhang et~al.(2021)Zhang, Li, and Li}]{zhang-etal-2021-ambert}
Xinsong Zhang, Pengshuai Li, and Hang Li. 2021.
\newblock {AMBERT}: A pre-trained language model with multi-grained
  tokenization.
\newblock In \emph{Findings of the ACL-IJCNLP 2021}, pages 421--435.
  Association for Computational Linguistics.

\bibitem[{Zhang and Yang(2018)}]{zhang-yang-2018-chinese}
Yue Zhang and Jie Yang. 2018.
\newblock {C}hinese {NER} using lattice {LSTM}.
\newblock In \emph{Proceedings of the 56th ACL}, pages 1554--1564. Association
  for Computational Linguistics.

\end{thebibliography}
\bibliographystyle{acl_natbib}




\end{CJK}
\end{document}